\documentclass[final]{cvpr}
\pdfoutput=1
\usepackage{times}
\usepackage{epsfig}
\usepackage{graphicx}
\usepackage{amsmath}
\usepackage{amssymb}

\usepackage{cite}

\usepackage{makecell}

\usepackage{microtype,graphicx,algorithm,algorithmic,multirow,booktabs,color} 
\usepackage[pagebackref=true,breaklinks=true,colorlinks,bookmarks=false]{hyperref}

\begin{document}

\title{Synergy Between Semantic Segmentation and Image Denoising\\ via Alternate Boosting}

\author{Shunxin Xu \quad  Ke Sun \quad Dong Liu \quad Zhiwei Xiong \quad Zheng-Jun Zha\\
University of Science and Technology of China\\

}

\maketitle

\begin{abstract}
The capability of image semantic segmentation may be deteriorated due to noisy input image, where image denoising prior to segmentation helps. Both image denoising and semantic segmentation have been developed significantly with the advance of deep learning. Thus, we are interested in the synergy between them by using a holistic deep model. We observe that not only denoising helps combat the drop of segmentation accuracy due to noise, but also pixel-wise semantic information boosts the capability of denoising. We then propose a boosting network to perform denoising and segmentation alternately. The proposed network is composed of multiple segmentation and denoising blocks (SDBs), each of which estimates semantic map then uses the map to regularize denoising. Experimental results show that the denoised image quality is improved substantially and the segmentation accuracy is improved to close to that of clean images. Our code and models will be made publicly available.
\end{abstract}

\section{Introduction}
Image semantic segmentation is a fundamental task in image understanding. The goal of semantic segmentation is to assign a category label for each pixel in an input image, which can be seen as pixel-level classification. Semantic segmentation has been actively studied in recent years because of its broad applications, such as augmented reality, autonomous driving, and satellite image analysis. Recent methods~\cite{chen2017deeplab, lin2017refinenet, ronneberger2015u, yang2018denseaspp} were inspired by the groundbreaking work named Fully Convolutional Network (FCN)~\cite{long2015fully}. FCN discarded fully connected layers, which were adopted in previous approaches, and was able to deal with arbitrary resolution.

In the real world, the capability of semantic segmentation may be deteriorated due to the noisy input image. As we have observed in our experiments on Cityscapes~\cite{cordts2016cityscapes} dataset, the segmentation accuracy obtained on noisy images is much lower than that on clean images by as high as 10 percent in mIoU, when the noise level (standard deviation of the additive Gaussian noise) is 50. Performing denoising prior to segmentation is a straightforward idea, and is verified effective in our experiments. Note that image denoising, as one of the most classic problems in image processing, has been enhanced greatly by the powerful convolutional neural network (CNN)~\cite{zhang2017beyond, mao2016image, zhang2018ffdnet} in recent years. Besides, more and more research efforts have been put on real-world noisy images instead of simulated (e.g. Gaussian) noise~\cite{anwar2019real, chen2019real, guo2019toward}. However, almost all of the existing denoising works pursue visual quality of denoised image, with little concern of the utility of denoising for downstream tasks (like segmentation).

We go one step further to ask how the capability of segmentation and denoising can be further improved. We suppose that semantic segmentation results may be helpful for denoising, too. We are motivated by a well-known denoising method, BM3D~\cite{dabov2007image}, where non-local correlation is used to enhance denoising ability via collaborative processing of similar image content. Thus, given a semantic segmentation map, the image content similarity may be better identified, and the non-local correlation may be better exploited.

With the aforementioned twofold motivations, we are interested in the synergy between image denoising and semantic segmentation by using a holistic deep model. Since the denoised image can be better segmented, and segmentation result can assist in denoising, we propose a boosting method to perform the two tasks. Specifically, we propose a segmentation and denoising alternate boosting network (SDABN). SDABN consists of multiple segmentation and denoising blocks (SDBs), each of which estimates a segmentation probability map from a noisy image and then uses the map to regularize image denoising. The output of one SDB is taken as the input to the next SDB, making a cascade that resembles boosting. Note that boosting has been proposed to improve image denoising~\cite{charest2006general,chen2019real} and classification~\cite{han2016incremental,moghimi2016boosted}. But our network is different as we perform two tasks--image denoising and semantic segmentation--alternately. We verify that the alternate boosting idea improves the performance of both denoising and segmentation.

In summary, we have made the following contributions in this paper:
\begin{itemize}
	\item We investigate the problem of semantic segmentation on the noisy image, and we propose to use denoising to combat the drop of segmentation accuracy due to noise. Noisy image is a common factor of performance drop. Thus, our method is useful for real-world image segmentation.
	\item We propose to use semantic segmentation results as additional conditions to regularize image denoising network. We verify that the additional information helps improve the capability of image denoising. Not only the visual quality of the denoised image is improved, but also the segmentation-aware denoised image can be better segmented further.
	\item We propose an alternate boosting network to exploit the synergy between denoising and segmentation. Our experimental results on the Cityscapes~\cite{cordts2016cityscapes} and OutdoorSeg~\cite{wang2018recovering} datasets with different kinds of noise demonstrate the superior performance of our method.
\end{itemize}

\section{Related Work}
\subsection{Semantic Segmentation}
Recently, researchers have been showing great interest in semantic segmentation. Fully Convolutional Network (FCN)~\cite{long2015fully} first discarded the fully connected layers and adopted the convolution layers throughout. Because of this, FCN can deal with arbitrary resolution images and becomes the most popular network architecture in semantic segmentation. 
To increase the receptive field, some works~\cite{yu2015multi,chen2017deeplab} adopted the dilated convolution layers in FCN. Later, most network structure for semantic segmentation can be regarded as encoder-decoder~\cite{badrinarayanan2017segnet,lin2017refinenet,zhao2017pyramid}. Encoder reduces the resolution of the feature maps, which can enlarge the receptive field, while decoder upsamples the feature maps to restore the resolution. The dilemma of these networks is that downsampling in encoder loses the details of the high-resolution feature maps. To address this problem, U-net~\cite{ronneberger2015u} introduced skip-connections to provide the decoder with feature maps of the encoder with the same resolution. Furthermore, HRNet~\cite{wang2020deep} maintained the high-resolution feature maps with a plain convolutional network and fused different resolution feature maps with concatenation. 
\subsection{Image Denoising}
Image denoising is the most basic problem in image processing and has been studied for several decades. Dictionary-based~\cite{dong2011sparsity,mairal2009non,elad2006image} methods solved the denoising problem by learning the sparse dictionary from clean images and coding the noisy images with the dictionary. BM3D~\cite{dabov2007image}, grouping similar patches for each location and filtering the groups collaboratively, is one of the mile-stones in denoising algorithm. Later, due to the development of deep learning, image denoising has made great progress in recent years. Zhang \emph{et al.}~\cite{zhang2017beyond} proposed DnCNN, which adopted residual learning~\cite{he2016deep} rather than learning the pair relation between noisy and cleaning images directly. Batch normalization~\cite{ioffe2015batch} also played a key role in their network. After that, some works~\cite{anwar2019real,chen2019real} proposed different networks for denoising, which concentrated on the structure of networks. Besides,~\cite{lehtinen2018noise2noise,krull2019noise2void} proposed schemes which can train networks with noisy images only, and~\cite{ulyanov2018deep} provided a method of training networks with a single noisy image. Boosting, an algorithm for improving the performance of various tasks by cascading the same models, has been adopted in image denoising not only in the traditional way~\cite{charest2006general,talebi2012saif} but also with CNN-based approach~\cite{chen2019real}.

\begin{figure*}[!t]
	\centering
	\includegraphics[scale=0.66]{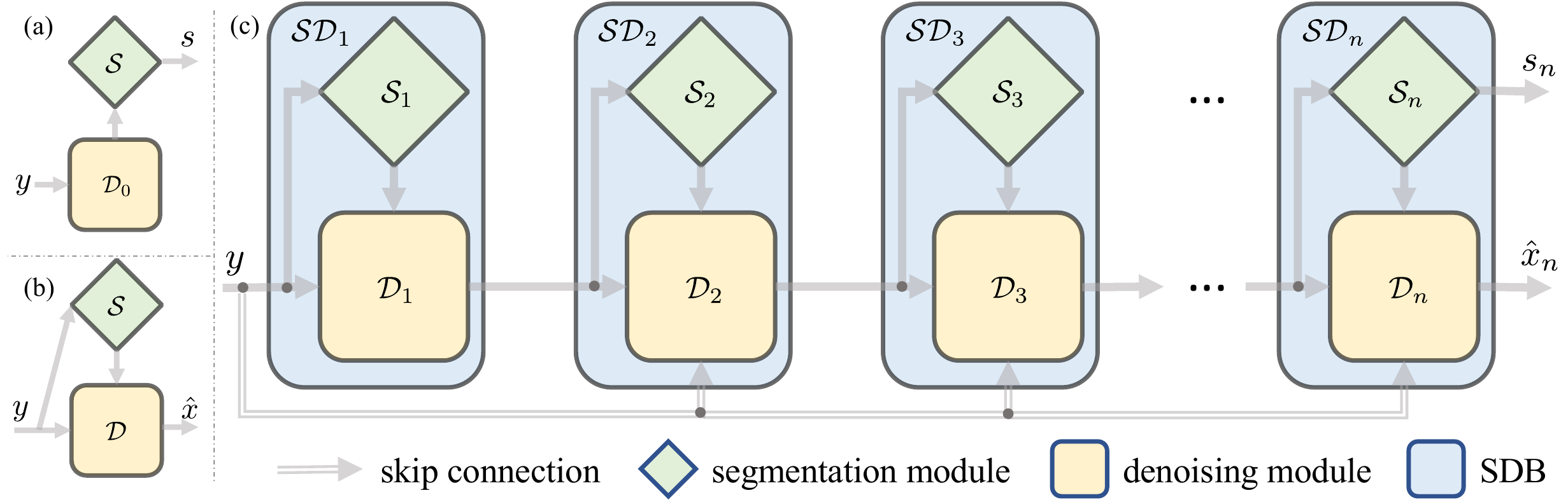}
	\caption{(a) Denoising followed by segmentation; (b) Segmentation followed by denoising; (c) The proposed segmentation and denoising alternate boosting network (SDABN), which is the concatenation of a series of segmentation and denoising blocks (SDBs). $y$, $s$, $\hat{x}$ stand for noisy image, segmentation (probability) map, and denoised image respectively.}
	\label{SDABN}
\end{figure*}

\subsection{Synergy Between High-level and Low-level Vision Tasks}
The high-level tasks, such as classification and semantic segmentation, are also regarded as image understanding, aiming to explore the semantic information of images. The low-level tasks, such as image denoising and super-resolution, are also seen as image processing, targeting the restoration of degraded images. There were already some researches on the synergy of these two task categories.~\cite{anwar2017category, remez2018class} brought the class prior to image denoising and needed to train a specific denoiser for each image class. In video deblurring,~\cite{ren2017video} estimated the segmentation probability map first and finished deblurring via the segmentation prior. To improve classification,~\cite{sharma2018classification} enhanced the target image and~\cite{zhang2019two} increased the resolution of the input image, and~\cite{wang2020deep} restored the degraded images to clean images in the feature domain of a classifier.  These works above focused on the performance of only one task.

In~\cite{zhang2018joint, wang2020dual}, the authors adopted multi-task frameworks to combine semantic segmentation and other tasks (e.g. depth estimation and super-resolution), but not denoising. Liu \emph{et al.}~\cite{liu2018image} took into account denoising and segmentation in one framework, which improved the segmentation performance of the denoising results by adding a segmentation loss as regularization to MSE loss, and we will detail more comparisons with this most related work in Sec.~\ref{sec_comparison}.

\begin{figure}[!h]
	\centering
	\includegraphics[scale=0.33]{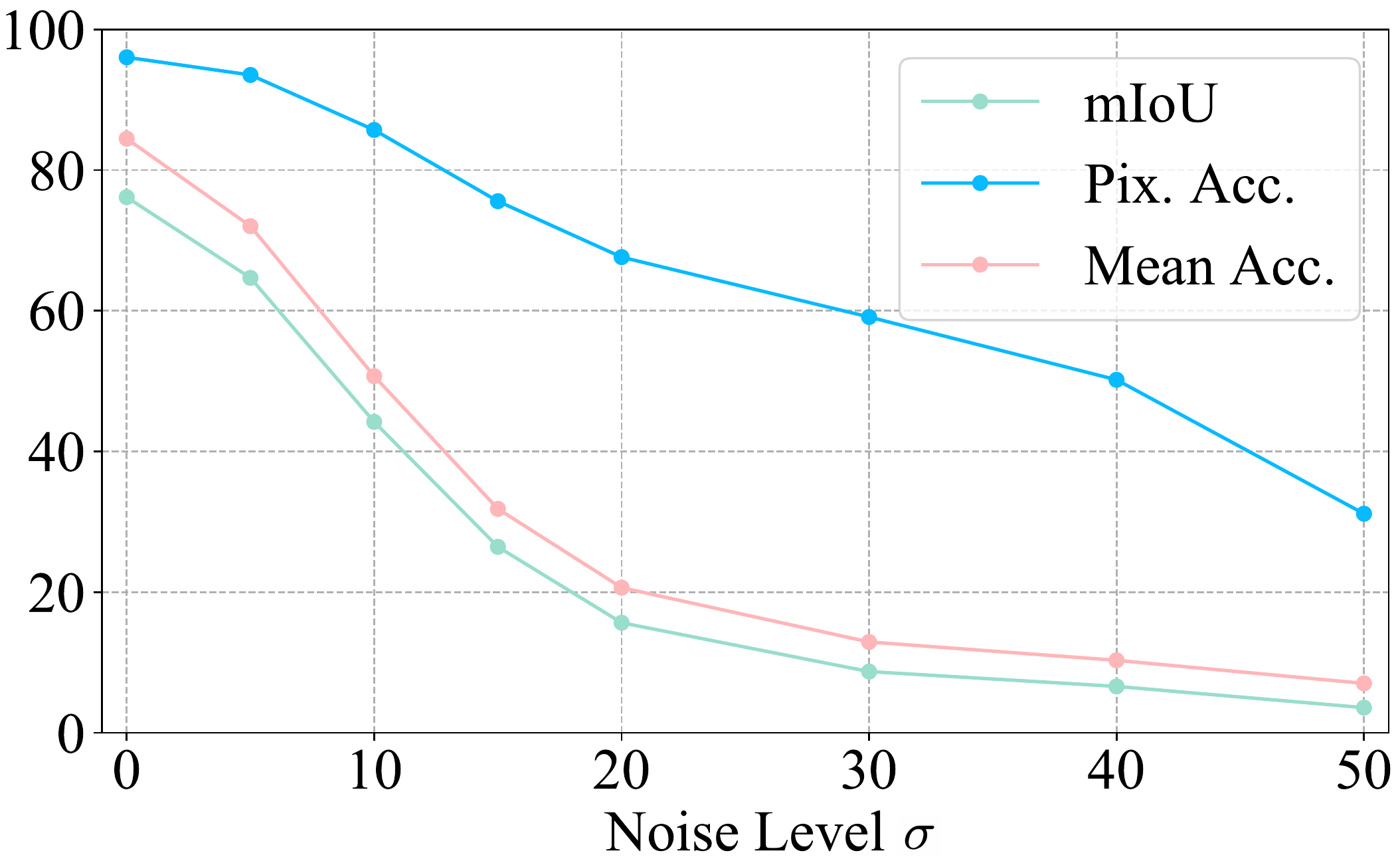}
	\caption{Segmentation results of $\mathcal{S}$ trained on clean images and tested on noisy images at different noise levels of Gaussian noise (noise level refers to the standard deviation of the added noise).}
	\label{nl}
\end{figure}

\begin{table}[!t]
	\begin{center}
		\begin{tabular}{c|c}
			\hline
			Denoising for seg. &  mIoU / Pix. Acc. / Mean Acc.  \\
			\hline
			$\mathcal{S}$ on clean images  &76.2 / 96.0 / 84.5  \\
			$\mathcal{S}$ on noisy images &63.9 / 93.4 / 74.0\\
			$\mathcal{S}$ on denoised images &64.7 / 93.8 / 74.7\\
			\hline\hline
			Seg. for denoising & PSNR / SSIM \\
			\hline
			$\mathcal{D}_0$&34.16 / 0.9082 \\
			$\mathcal{S}\mathcal{D}$&\textbf{34.23} / \textbf{0.9100}\\
			$Img\mathcal{D}$&34.09 / 0.9072\\
			\hline
		\end{tabular}
	\end{center}
	\caption{Results of the explorative study (c.f. Sec.~\ref{sec_observation}) with noise level $\sigma = 50$. $\mathcal{D}_{0}$ is direct denoising of noisy images. $\mathcal{S}\mathcal{D}$ is segmentation followed by denoising (Fig.~\ref{SDABN}(b)), where the denoiser structure is Fig.~\ref{ddfn_sft}(a). $Img\mathcal{D}$ is using the denoiser structure in Fig.~\ref{ddfn_sft}(a) but replacing the input segmentation map with noisy images.}
	\label{observation}
\end{table}

\section{Method}

\begin{figure*}[!t]
	\centering
	\includegraphics[scale=0.71]{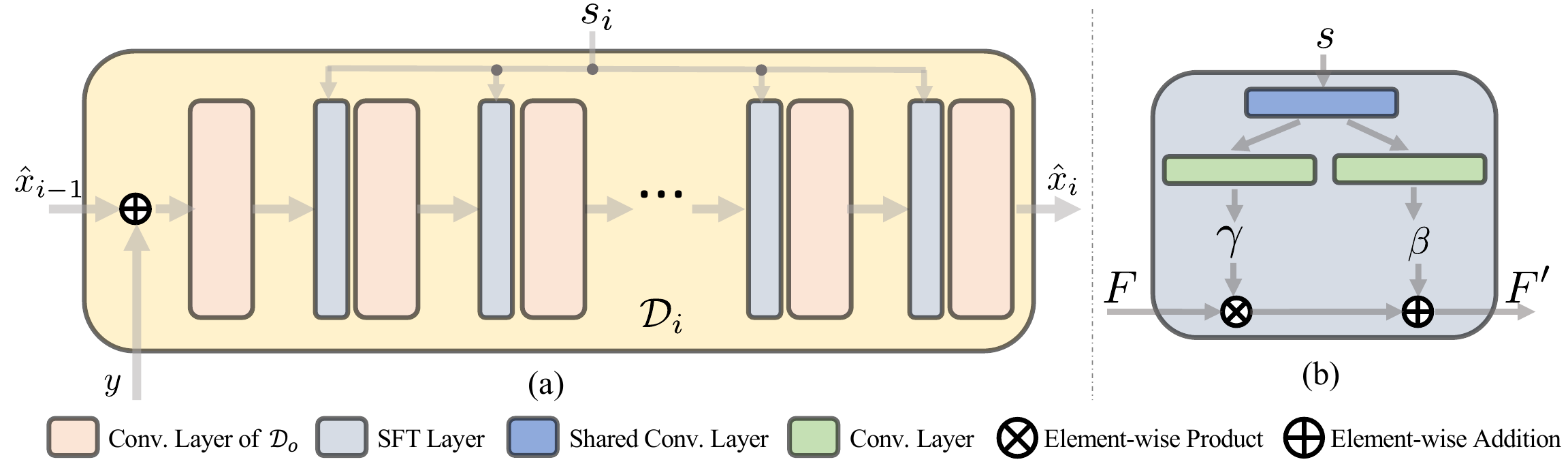}
	\caption{(a) The structure of denoising module $\mathcal{D}_i$ in Fig.~\ref{SDABN}, note that $\hat{x}_{0}=0$; (b) The structure of the spatial feature transform (SFT) layer in (a).}
	\label{ddfn_sft}
\end{figure*}

\subsection{Two Observations}
\label{sec_observation}
In the beginning, we present two observations about the synergy between image denoising and semantic segmentation. The observations are based on Fig.~\ref{nl} and Table~\ref{observation}, where the observations are on Cityscapes~\cite{cordts2016cityscapes} and noisy data is simulated with additive white Gaussian noise (AWGN). 

From the results illustrated in Fig.~\ref{nl}, we find the performance of a segmentation network $\mathcal{S}$, trained on clean images, will drop severely when tested on noisy images. Then we continue to explore the situation where $\mathcal{S}$ is with the consistency between training and testing. For example, when we train a model with clean (or noisy, or denoised) images, we test it on clean (or noisy, or denoised) images, too.

\underline{\textit{\textbf{Denoising helps segmentation.}}} We train a segmentation network $\mathcal{S}$ on clean images, and the testing mIoU / pixel accuracy (Pix. Acc.) / mean accuracy  (Mean Acc.) is $76.2 / 96.0 / 84.5$. However, if we replace the training data with the corresponding noisy images and train a new segmentation module, its testing result drops to $63.9 / 93.4 / 74.0$. Because the denoising method can reduce the noise, it is intuitive to adopt a denoiser $\mathcal{D}_0$ before semantic segmentation to make up the performance drop from clean images to noisy images. As shown in Fig.~\ref{SDABN}(a), we train one denoising network $\mathcal{D}_0$ on Cityscapes and train another segmentation network $\mathcal{S}$ on images denoised by that denoiser. In this situation, the testing result increases to $64.7 / 93.8 / 74.7$. The comparisons present that noise has negative effects on semantic segmentation and denoising can reduce such effects. These results are reasonable because noise disturbs the intensity of each pixel, which also damages the semantic information of the image and worsens the performance of segmentation methods. Therefore, denoising, as a technique for noise reduction, can help the segmentation module on noisy images.

\underline{\textit{\textbf{Segmentation helps denoising.}}} Testing PSNR/SSIM of the denoiser $\mathcal{D}_0$ mentioned just now is $34.16/0.9082$. Then we suppose that semantic segmentation results may be helpful for denoising, too. 
We plug a new segmentation module $\mathcal{S}$ for one denoising network to construct a segmentation-aware denoiser, and this combination is as our basic module SDB, denoted as $\mathcal{S}\mathcal{D}$. The denoiser accepts a segmentation condition estimated by $\mathcal{S}$ and is denoted as $\mathcal{D}$, which is shown in Fig.~\ref{SDABN}(b), and its denoising result becomes to $34.23/0.9100$. (Layers of the denoising network are normalized according to the segmentation condition, and more details will be introduced in Sec.~\ref{sec_SDB}.) However, more learnable parameters are brought to $\mathcal{D}$ when it accepts an additional condition, compared with $\mathcal{D}_0$. To validate where the performance gain originates from, we replace the segmentation condition with the noisy images to train (and test) $\mathcal{D}$, denoted as $Img\mathcal{D}$, which means the noisy images are inputted as not only denoising target but also condition. $Img\mathcal{D}$ is with the same parameter number as $\mathcal{S}\mathcal{D}$ in its denoising module but gets the worst denoising results. It indicates that the gain of $\mathcal{S}\mathcal{D}$ comes from the segmentation prior rather than the additional learnable parameters. Consequently, the experiments verify that the estimated segmentation probability map can indeed improve the performance of the denoiser. 

\subsection{Segmentation and Denoising Block (SDB)}
\label{sec_SDB}

For the denoising task, we define the clean image as $x \in \mathbb{R}^{H\times W\times C}$, where $C$ denotes the channel number of the image. Unless noted otherwise, we adopt RGB images in this paper, so C is equal to 3. Then the noisy image $y$ can be formed as $y=\mathcal{V}(x)$, where $\mathcal{V}$ stands for the mapping from the clean image to the noisy image. Finally, with a denoising method $\mathcal{D}_{0}$, the denoising result $\hat{x}$ can be formulated as $\hat{x}=\mathcal{D}_0(y)$.

In SDB, the input $y$ first goes through a semantic segmentation module denoted as $\mathcal{S}$, and the estimated segmentation probability maps can be formed by $s=\mathcal{S}(y)$, where $s \in \mathbb{R}^{H\times W\times N}$ and $N$ is the number of the categories in the dataset. The function of SDB can be formulated as:
\setlength{\abovedisplayskip}{5pt}
\setlength{\belowdisplayskip}{5pt}
\begin{equation}
	\label{SDB}
	\left\{
	\begin{array}{lr}
		s=\mathcal{S}(y),   \\
		\hat{x}=\mathcal{D}(y, s). \\
	\end{array}
	\right.
\end{equation}

To utilize the estimated segmentation probability maps $s$ in denoising, we insert Spatial Feature Transform (SFT) layers~\cite{wang2018recovering} to the backbone $\mathcal{D}_0$ as our segmentation-aware denoiser $\mathcal{D}$ (see Fig.~\ref{ddfn_sft}(a)). The SFT layer can be regarded as a conditional normalization layer, which transforms some feature maps according to the given condition $s$. We illustrate SFT in Fig.~\ref{ddfn_sft}(b). Specifically, there are two inputs: One is the feature maps $F$ from $\mathcal{D}_0$, and the other is the segmentation probability maps $s$ as the condition. The outputs of SFT are the normalized feature maps $F'=\gamma\odot F+\beta$, where $(\gamma,\beta)$ are the coefficients of linear transformation mapped from $s$ through convolutional layers, and $\odot$ denotes the element-wise multiplication. Following~\cite{wang2018recovering}, we apply SFT to the layers of $\mathcal{D}_0$ except the first one.

The idea of applying SFT to utilize the segmentation maps is inspired by the denoising method BM3D~\cite{dabov2007image}, where non-local correlation is used to enhance denoising ability via collaborative processing of similar image content. In SDB, pixels are normalized by SFT based on the segmentation maps, which provide non-local consistency of the category information.

In implement, we adopt HRNetV2-W18-Small-v2~\cite{wang2020deep} and Dilated Dense Fusion Network (DDFN)~\cite{chen2019real} with SFT~\cite{wang2018recovering} as the basic segmentation and denoising module in our network, respectively, which both achieve the state-of-the-art in their tasks. \textit{{Note that the concentration of this paper is the exploration of the synergy rather than the network design.}} These two modules are chosen to realize our explorations, which can be replaced by any two networks for segmentation and denoising, respectively.

\subsection{Segmentation and Denoising Alternate Boosting Network (SDABN)}
\label{sec_SDABN}

Because the denoised image can be better segmented and the segmentation result can assist in denoising, we can use outputs of the segmentation-aware denoiser and the denoiser-assisted segmentation to enhance the further segmentation and denoising modules, respectively.  Inspired by these, we construct our segmentation and denoising alternate boosting network (SDABN) and show it in Fig.~\ref{SDABN}(c), which is a cascade of several basic SDBs. Based on (\ref{SDB}), the formula of SDABN can be formulated iteratively:
\setlength{\abovedisplayskip}{5pt}
\setlength{\belowdisplayskip}{5pt}
\begin{equation}
	\label{all}
	\left\{
	\begin{array}{lr}
		s_{i}=\mathcal{S}_{i}(\hat{x}_{i-1}), \\
		\hat{x}_{i}=\mathcal{D}_{i}(\hat{x}_{i-1}, s_{i}, y), \\
		s_{1}=\mathcal{S}_{1}(y), \\
		\hat{x}_{1}=\mathcal{D}_{1}(y, s_{1}),
		i \in \{2,3,...,n\},
	\end{array}
	\right.
\end{equation}
where $i$ indicates the different SDB and $n$ stands for the number of the blocks in total.

In SDABN,  each $\mathcal{S}_{i}$ will be helped by the former segmentation-aware denoiser $\mathcal{SD}_{i-1}$ and also estimates semantic maps to assist the paired denoiser $\mathcal{D}_{i}$, here $i\geq 2$. Ignoring the segmentation part of each SDB temporarily, our SDABN can be regarded as a cascaded denoiser, whose performance is boosted with the basic unit number increasing~\cite{chen2019real}. In Fig.~\ref{SDABN}(c) and (\ref{all}), there are skip connections between the noisy image and inputs of each denoising block. Chen \emph{et al.}~\cite{chen2019real} claimed that these connections are helpful for the boosting of image denoising.

As a result, the denoising capacity in SDABN is improved by not only segmentation prior but also the former denoising results. As for segmentation, its performance is improved by denoising, which has been shown in Sec.~\ref{sec_observation}. Besides, we will verify that there is also boosting between segmentation modules in Sec.~\ref{salient}, which is bridged by the middle denoiser.

\subsection{Training Strategy}
\label{sec_strategy}
We train SDABN progressively in the order of $\mathcal{S}_1\rightarrow\mathcal{D}_1\rightarrow\mathcal{S}_{2}\rightarrow\mathcal{D}_{2}\rightarrow...$ All the trained modules are fixed when we train the next one.  For training the denoising block $\mathcal{D}_i$, the loss function $\mathcal{L_{D}}_{i}$ is the mean square error (MSE) between the output of the current block $\hat{x}_{i}$ and the clean image $x$. The loss function for semantic segmentation block $\mathcal{S}_i$ is the cross-entropy loss $\mathcal{L_{S}}_{i}$ between the estimated probability maps $s_i$ and the segmentation label $l$. The training strategy is shown in Algorithm~\ref{alg}.

\section{Experiments}

In this section, $\mathcal{S}_1\mathcal{D}_1\mathcal{S}_2\mathcal{D}_2...\mathcal{S}_n\mathcal{D}_n$ denotes a network composed of $n$ SDBs, and $\mathcal{S}_1\mathcal{D}_1\mathcal{S}_2\mathcal{D}_2...\mathcal{S}_n$ stands for a $\mathcal{S}_1\mathcal{D}_1\mathcal{S}_2\mathcal{D}_2...\mathcal{S}_n\mathcal{D}_n$ discarding the last denoising module. DDFN$\times n$ represents a cascade of $n$ DDFNs~\cite{chen2019real}. 

\begin{figure*}[!h]
	\centering
	\includegraphics[scale=0.13]{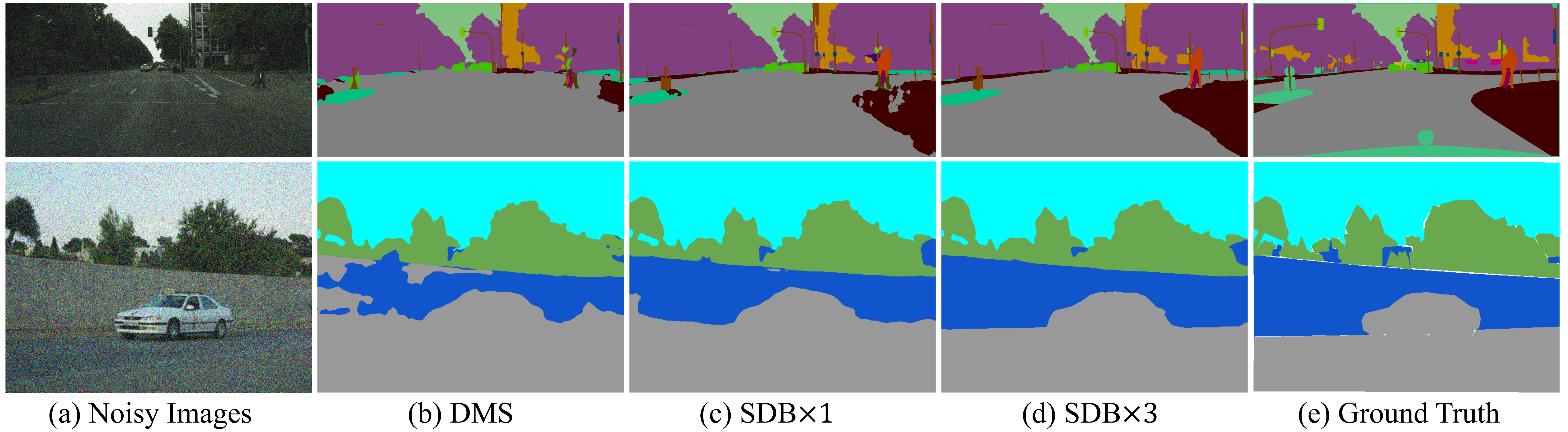}
	\caption{Segmentation results of DMS~\cite{liu2018image} and our method when noise level $\sigma=50$. Top row: one image from the Cityscapes set; Bottom row: one image from the OutdoorSeg set.}
	\label{visual_seg}
\end{figure*}
\begin{figure*}[!h]
	\centering
	\includegraphics[scale=0.13]{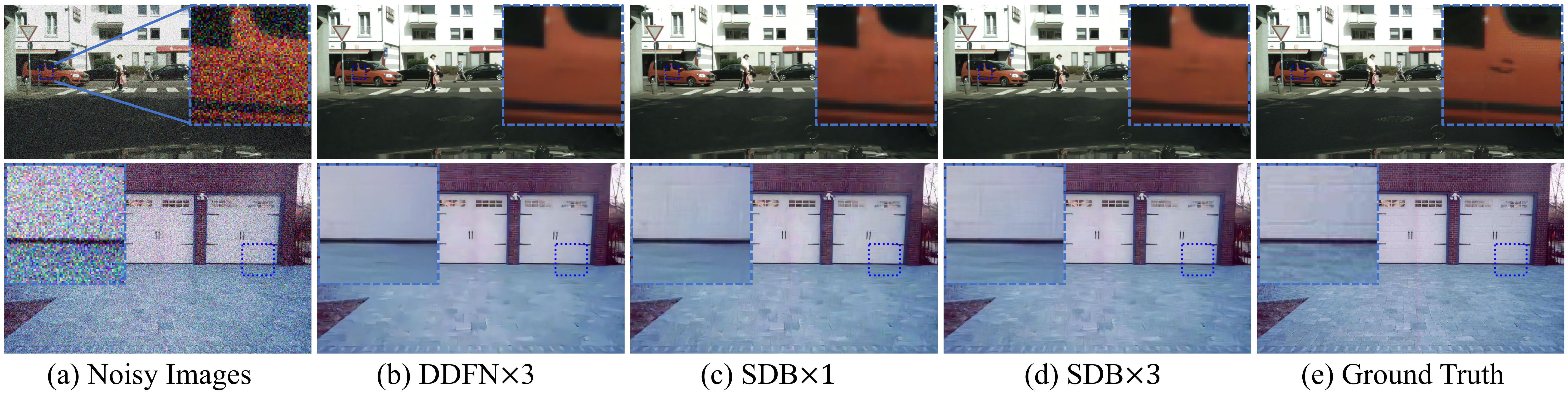}
	\caption{Denoising results of DDFN$\times3$~\cite{chen2019real} and our method when noise level $\sigma=50$. Top row: one image from the Cityscapes set; Bottom row: one image from the OutdoorSeg set.}
	\label{visual_dn}
\end{figure*}

\subsection{Experiment Settings}
It is very difficult to conduct experiments on real noisy images due to the lack of such data with both real-world noise and segmentation labels, so we use AWGN to simulate the noisy images. As a supplement, we will validate the generalization ability in Sec.~\ref{sec_generalization}.

We train our network on two datasets: Cityscapes~\cite{cordts2016cityscapes} and OutdoorSeg~\cite{wang2018recovering}. For Cityscapes, we follow the original splits: The dataset is divided into a training set with $2975$ images and a validation (val) set with $500$ images. OutdoorSeg is a merged dataset for outdoor scene segmentation, which collects $9900$ outdoor images from ADE dataset~\cite{zhou2017scene}, COCO dataset~\cite{lin2014microsoft} and Flickr website. In this paper, we divide these $9900$ images into $8800$ and $1100$ as training and testing sets respectively. More implementation details are listed in the supplementary material.

\renewcommand{\algorithmicrequire}{\textbf{Input:}}
\begin{algorithm}[!t]
	\small
	\caption{Training SDABN}  
	\begin{algorithmic} [1]
		\REQUIRE noisy images $y$, clean images $x$ and segmentation labels $l$
		\FOR{ $i = 1,2,3,...,n$}
		\IF {$i=1$}
		\STATE Initialize the parameters of $\mathcal{S}_1 $ by the model trained on clean images;
		\ELSE 
		\STATE Initialize the parameters of $\mathcal{S}_i $ by those of $\mathcal{S}_{i-1} $;
		\ENDIF
		\WHILE  {not converge}
		\STATE Sample minibatch of noisy images $y$ and paired segmentation labels $l$ from dataset;
		\STATE Optimize the parameters of $\mathcal{S}_i $ by SGD on $\mathcal{L_{S}}_{i}$;
		\ENDWHILE
		\STATE Fix the parameters of segmentation block $\mathcal{S}_i $;
		\WHILE {not converge}
		\STATE Sample minibatch of noisy images $y$ and corresponding clean images $x$ from dataset;
		\STATE Optimize the parameters of $\mathcal{D}_i $ by Adam on $\mathcal{L_{D}}_{i}$;
		\ENDWHILE
		\STATE Fix the parameters of denoising block $\mathcal{D}_i $;
		\ENDFOR
	\end{algorithmic}  
	\label{alg}
\end{algorithm}

\begin{table}[!h]
	\footnotesize
	\begin{center}
		\begin{tabular}{|c|c|c|c|c|c|}
			\cline{1-6}
			\multicolumn{6}{|c|}{Cityscapes} \\
			\cline{1-6}
			\thead{$\sigma$} &\thead{Metric}&\thead{BM3D~\cite{dabov2007image} \\+Seg.}&\thead{DDFN~\cite{chen2019real}\\+Seg.}&\thead{DMS~\cite{liu2018image}}&\thead{SDB}\\
			\cline{1-6}
			\multirow{5}*{$50$} &PSNR&33.68&34.16&33.74&\textbf{34.23}\\
			~&SSIM&0.8953&0.9082&0.9005&\textbf{0.9100}\\
			\cline{2-6}
			~&mIoU&39.2&40.5&51.9&\textbf{63.9}\\
			~&Pix. Acc.&69.6&72.5&88.6&\textbf{93.4}\\
			~&Mean Acc.&56.9&59.2&61.4&\textbf{74.0}\\
			\cline{1-6}
			\multirow{5}*{$30$} &PSNR&35.78&36.18&35.79&\textbf{36.27}\\
			~&SSIM&0.9224&0.9315&0.9260&\textbf{0.9328}\\
			\cline{2-6}
			~&mIoU&55.5&48.1&59.5&\textbf{67.1}\\
			~&Pix. Acc.&84.2&77.7&91.8&\textbf{94.4}\\
			~&Mean Acc.&68.5&67.0&68.6&\textbf{76.1}\\
			\cline{1-6}
			\multirow{5}*{$10$} &PSNR&40.59&40.75&40.12&\textbf{40.83}\\
			~&SSIM&0.9651&0.9668&0.9622&\textbf{0.9671}\\
			\cline{2-6}
			~&mIoU&71.8&69.5&69.4&\textbf{73.4}\\
			~&Pix. Acc.&95.0&94.4&94.7&\textbf{95.5}\\
			~&Mean Acc.&81.4&80.6&77.9&\textbf{82.3}\\
			\cline{1-6}	
			\multicolumn{6}{|c|}{OutdoorSeg} \\
			\cline{1-6}
			\thead{$\sigma$} &\thead{Metric}&\thead{BM3D~\cite{dabov2007image} \\+Seg.}&\thead{DDFN~\cite{chen2019real}\\+Seg.}&\thead{DMS~\cite{liu2018image}}&\thead{SDB}\\
			\cline{1-6}
			\multirow{5}*{$50$} &PSNR&27.52&27.97&27.89&\textbf{28.01}\\
			~&SSIM&0.7875&0.8106&0.8063&\textbf{0.8115}\\
			\cline{2-6}
			~&mIoU&74.1&72.8&75.0&\textbf{75.2}\\
			~&Pix. Acc.&87.4&86.3&87.9&\textbf{88.3}\\
			~&Mean Acc.&84.7&82.4&85.1&\textbf{85.8}\\
			\cline{1-6}
			\multirow{5}*{$25$} &PSNR&31.02&31.35&31.22&\textbf{31.39}\\
			~&SSIM&0.8880&0.8966&0.8925&\textbf{0.8973}\\
			\cline{2-6}
			~&mIoU&76.9&76.4&\textbf{77.5}&77.4\\
			~&Pix. Acc.&88.9&88.3&89.2&\textbf{89.3}\\
			~&Mean Acc.&86.0&85.2&86.7&\textbf{86.9}\\
			\cline{1-6}
		\end{tabular}
	\end{center}
	\vspace{-0.22cm}
	\caption{Quantitative comparison results of denoising and segmentation. BM3D~\cite{dabov2007image}+Seg. and DDFN~\cite{chen2019real}+Seg. stand for different denoisers~\cite{dabov2007image,chen2019real} followed by the segmentation module trained on clean images, respectively. DMS~\cite{liu2018image} is a joint denoising-segmentation method. }
	\label{comparison}
\end{table}

\subsection{Comparison with the Previous Method}
\label{sec_comparison}

Liu \emph{et al.} propose a framework, named Denoising Meeting Segmentation (DMS)~\cite{liu2018image}, taking into account denoising and segmentation. They add an extra segmentation loss to the original denoising loss (MSE) with a fixed segmentation module, which is pretrained on clean images, and a denoising module is updated according to the joint loss. The roles of the segmentation module and loss are similar to that of VGG~\cite{simonyan2014very} and the perceptual loss~\cite{johnson2016perceptual} in super-resolution, respectively. 
The conceptual differences of our method and DMS~\cite{liu2018image} are twofold. First,  we use an alternate boosting framework whereas DMS uses a multi-task one. Second, we optimize both segmentation and denoising to adapt to the noisy images, but their segmentation module is fixed, only the denoising module is trained. 

The comparison results are in Table~\ref{comparison}. Note that we reproduce DMS by applying HRNetV2-W18-Small-v2~\cite{wang2020deep} and DDFN~\cite{chen2019real} as the segmentation and denoising module, respectively, which are the same as those in our method. To be fair to DMS, only one SDB is adopted to compare with it. Besides, we provide the results of BM3D~\cite{dabov2007image} and DDFN~\cite{chen2019real}  followed by a segmentation module trained on clean images as another two baselines and denote them as BM3D~\cite{dabov2007image}+Seg. and DDFN~\cite{chen2019real}+Seg., respectively.

From Table~\ref{comparison}, we can notice that the denoising results of DMS~\cite{liu2018image} are even worse than our common denoising backbone DDFN~\cite{chen2019real}. 
Liu \emph{et al.}~\cite{liu2018image} explained that the extra segmentation loss, regarded as a regularization term, disturbs the original optimization on MSE. But in our method, the denoising module is only trained with MSE and is always superior to BM3D, DDFN, and DMS. As for segmentation, our method is also the best.  The segmentation module in the three other methods is fixed after pretrained on clean images. Even though DMS adopts the segmentation loss as guidance, our method still outperforms it because we optimize the segmentation module for its inputs. From the comparison between BM3D~\cite{dabov2007image}+Seg. and DDFN~\cite{chen2019real}+Seg., we can observe that, for different denoisers, better denoising performance does not always lead to better segmentation results when the segmentation network is fixed.

\begin{table*}[!h]
	\begin{center}
		\begin{tabular}{c|c|c|cc|cc|ccc}
			\cline{1-10}
			\multirow{2}*{Dataset}&\multirow{2}*{$\sigma$}&\multirow{2}*{\#Units}  &\multicolumn{2}{c|}{DDFN~\cite{chen2019real}} & \multicolumn{5}{c}{SDB} \\
			\cline{4-10}
			~&~&~&PSNR&SSIM  & PSNR &SSIM& mIoU&Pix. Acc.&Mean Acc. \\
			\cline{1-10}
			\multirow{10}*{Cityscapes}
			~&\multirow{3}*{$50$} &$\times1$&34.16&0.9082&34.23&0.9100&63.9&93.41&74.0\\
			~&~&$\times2$&34.28 & 0.9099&34.34&0.9115&66.1&94.05&\textbf{76.3}\\
			~&~&$\times3$&34.34 & 0.9108&\textbf{34.42}&\textbf{0.9122}&\textbf{66.5}&\textbf{94.06}&75.6\\
			\cline{2-10}
			~&\multirow{3}*{$30$} &$\times1$&36.18&0.9315&36.27&0.9328&67.1&94.42&76.1\\
			~&~&$\times2$&36.28 & 0.9323&36.39&0.9340&68.5&94.77&77.8\\
			~&~&$\times3$&36.34 & 0.9330&\textbf{36.50}&\textbf{0.9351}&\textbf{69.0}&\textbf{94.80}&\textbf{78.6}\\
			\cline{2-10}
			~&\multirow{3}*{$10$} &$\times1$&40.75&0.9668&40.83&0.9671&73.4&95.47&82.3\\
			~&~&$\times2$&40.84 & 0.9671&40.94&0.9677&74.4&95.65&82.9\\
			~&~&$\times3$&40.88 & 0.9672&\textbf{41.06}&\textbf{0.9683}&\textbf{74.8}&\textbf{95.73}&\textbf{83.3}\\
			\cline{2-10}
			~&$0$ &-&-&-&-&-&76.2&96.0&84.5\\
			\cline{1-10}
			\multirow{7}*{OutdoorSeg}
			~&\multirow{3}*{$50$} &$\times1$&27.97&0.8106&28.01&0.8115&75.2&88.25&85.8\\
			~&~&$\times2$&28.06 & 0.8136&28.10&0.8152&76.4&88.85&86.0\\
			~&~&$\times3$&28.10 & 0.8147&\textbf{28.14}&\textbf{0.8159}&\textbf{76.7}&\textbf{89.11}&\textbf{86.5}\\
			\cline{2-10}
			~&\multirow{3}*{$25$} &$\times1$&31.35&0.8966&31.39&0.8973&77.4&89.32&86.9\\
			~&~&$\times2$&31.42 & 0.8988&31.47&0.8995&78.0&89.57&87.4\\
			~&~&$\times3$&31.43 & 0.8990&\textbf{31.50}&\textbf{0.9002}&\textbf{78.2}&\textbf{89.68}&\textbf{87.5}\\
			\cline{2-10}
			~&$0$ &-&-&-&-&-&78.9&90.0&87.9\\
			\cline{1-10}
		\end{tabular}
	\end{center}
	\caption{Quantitative results of boosting for~\cite{chen2019real} and our method. Note that the segmentation results of SDB$\times1$ are also those of the segmentation backbone trained on noisy images since our SDABN is beginning with segmentation, and the rows with $\sigma=0$ are the results of the segmentation backbone on clean images.}
	\label{quan_result}
\end{table*}

\subsection{Boosting Results of SDABN}
In this section, we present the semantic segmentation results of SDBs and the denoising results of SDBs and DDFNs~\cite{chen2019real}. The quantitative results on Cityscapes and OutdoorSeg are shown in Table~\ref{quan_result}. We adopt PSNR and SSIM as the metrics for denoising and compute mIoU, Pix. Acc. and Mean Acc. for segmentation. For these five metrics, larger is better. Note that the segmentation results of SDB$\times1$ are also those of the segmentation backbone trained on noisy images since SDABN is beginning with segmentation, and the rows with $\sigma=0$ are the results of the segmentation backbone trained on clean images.

We compare the denoising results of DDFNs~\cite{chen2019real} and SDBs because DDFN is the backbone of our denoiser. Both of them have different numbers of basic units, which are denoted as $\times1$, $\times2$, and $\times3$. From the results of PSNR/SSIM, we can observe that SDBs always have better performance, especially on the Cityscapes dataset.

From the results of DDFNs, we know that denoising is boosted with the unit number increasing. There is the same conclusion of SDBs if we consider SDABN as a cascade of segmentation-aware denoisers. Hence, the denoising capacity of SDABN is improved by not only segmentation prior but also the former denoising results.

Besides, the results in Table~\ref{quan_result} verify that the performance of segmentation on noisy images is negatively correlated to the noise intensity in two ways. First, if we only consider the single segmentation module with the different noise levels, we can find that the performance on the images with a lower noise level is closer to that on clean images. 
Second, because the denoisers provide cleaner and cleaner images as the number of units increases, the segmentation performance is boosted and closer to the results of clean images.

Additionally, the visual results of segmentation in Fig.~\ref{visual_seg} show that our method has better performance on the boundary than DMS~\cite{liu2018image} and refines the results with the unit number increasing. The visual results of denoising in Fig.~\ref{visual_dn} show that our method is with more details than DDFN~\cite{chen2019real}, which is also boosted with the unit number increasing. 
\textit{Note that the improvement of denoising reflected by visual results is more significant, compared with that of quantitative results.} More visual results are provided in Supplementary Material.

\subsection{Boosting Between Segmentation Modules}
\label{salient}

 Denoising can be boosted with the basic unit number increasing~\cite{chen2019real}. In this section, we will verify that there is also boosting between segmentation modules. The related results are on Cityscapes with noise level 50 and shown in Table~\ref{nece_s}. We list 4 different denoisers in this table, where $Gt\mathcal{D}_1$ stands for $\mathcal{D}_1$ trained (and tested) with ground truth of the segmentation as a condition, rather than the estimated one from $\mathcal{S}_1$. Correspondingly, there are 4 segmentation modules, denoted as $\mathcal{X}+\mathcal{S}$,  where $\mathcal{X}\in\{$DDFN$\times1, $DDFN$\times2, \mathcal{S}_1\mathcal{D}_1, Gt\mathcal{D}_1\}$, and the inputs of $\mathcal{X}+\mathcal{S}$ are from the denoiser $\mathcal{X}$.

From Table~\ref{nece_s}, we can observe that DDFN$\times2$ is better than $\mathcal{S}_1\mathcal{D}_1$ on the denoising metrics, but DDFN$\times2+\mathcal{S}$ is worse than $\mathcal{S}_1\mathcal{D}_1+\mathcal{S}$ on the segmentation metrics. This observation demonstrates that the denoising results of the segmentation-aware denoiser for the following segmentation module achieve higher segmentation accuracy, although its denoising performance is worse. We define that such denoised images are with salient semantic information, which means they can be recognized by another segmentation module with higher accuracy. As for DDFN$\times n$ and DDFN$\times n+\mathcal{S}$, the correlation between denoising and segmentation performance is positive. Hence, the property about salient semantic information comes from the segmentation-aware design, which is the only difference between DDFN$\times 1$ and $\mathcal{S}_1\mathcal{D}_1$.
Besides, better quality segmentation prior helps restore images with more salient semantic information, whose evidence is segmentation labels have the best quality and $Gt\mathcal{D}_1+\mathcal{S}$ achieves the best performance on segmentation.

So we claim that our segmentation-aware design is helpful for the later segmentation module. In other words, the segmentation module $\mathcal{S}_i$ is beneficial to the later $\mathcal{S}_{i+1}$. We regard it as the boosting between segmentation modules, which is bridged by the middle denoiser $\mathcal{D}_i$. 
\begin{table}[!t]
	\small
	\begin{center}
		\begin{tabular}{cccc}
			\hline
			Denoising & Seg.-Aware & PSNR &SSIM \\
			\hline
			DDFN$\times1$ &$\times$ & 34.157&0.9082  \\
			DDFN$\times2$&$\times$& \textbf{34.280}&0.9099  \\
			$\mathcal{S}_1\mathcal{D}_1$&  $\checkmark$& 34.231&\textbf{0.9100} \\
			\hline
			$Gt\mathcal{D}_1$& $\checkmark$& 34.234 &0.9099\\
			\hline\hline
			Segmentation & mIoU & Pix. Acc.&Mean Acc.\\
			\hline
			DDFN$\times1+\mathcal{S}$& 64.7 &93.8&74.7\\
			DDFN$\times2+\mathcal{S}$& 65.7 &93.9&75.6\\
			$\mathcal{S}_1\mathcal{D}_1+\mathcal{S}$ & \textbf{66.1}&\textbf{94.1}&\textbf{76.3}\\
			\hline
			$Gt\mathcal{D}_1+\mathcal{S}$ & 87.0&98.4&92.3\\
			\hline
		\end{tabular}
	\end{center}
	\caption{Ablation study results (c.f. Sec.~\ref{salient}). DDFN refers to~\cite{chen2019real}. $Gt\mathcal{D}_1$ is using ground truth of the segmentation map as the input to $\mathcal{D}_1$ to test the ``oracle" performance.}
	\label{nece_s}
\end{table}

\begin{table}[!t]
	\small
	\begin{center}
		\begin{tabular}{ccccc}
			\cline{1-5}
			Method &$\mathcal{S}_1$& $\mathcal{S}_1+$ & $\mathcal{S}_1\mathcal{D}_1\mathcal{S}_2$-joint & $\mathcal{S}_1\mathcal{D}_1\mathcal{S}_2$\\
			\midrule
			\#Params. &3.94M&8.00M &7.99M &  7.99M \\
			mIoU&63.9& 65.1 &  57.1  &   \textbf{66.1}\\
			Pix. Acc.&93.4& 93.8 &  92.0  &   \textbf{94.1}\\
			Mean Acc.&74.0& 75.4 &  66.8  &   \textbf{76.3}\\
			\cline{1-5}
		\end{tabular}
	\end{center}
	\caption{Ablation study results (c.f. Sec.~\ref{sec_necessity}). $\mathcal{S}_1+$ is a wider and deeper version of $\mathcal{S}_1$. $\mathcal{S}_1\mathcal{D}_1\mathcal{S}_2$-joint is a single segmentation network with the same structure as $\mathcal{S}_1\mathcal{D}_1\mathcal{S}_2$.}
	\label{nece_d}
	\vspace{-0.3cm}
\end{table}

 \subsection{Necessity of the Denoising in SDABN}
\label{sec_necessity}
In this section, we will verify the necessity of the segmentation-aware denoising for segmentation on noisy images. From Table~\ref{quan_result}, we know that the segmentation performance is boosted with the basic unit number increasing. But the learnable parameters of $\mathcal{S}_1\mathcal{D}_1\mathcal{S}_2$ are more than those of $\mathcal{S}_1$,  so we wonder whether the segmentation-aware denoising modules or the more parameters lead to these improvements.

In SDABN, $\mathcal{S}_1$ and $\mathcal{S}_2$ are trained with segmentation loss, and  $\mathcal{D}_1$ is trained with denoising loss.  $\mathcal{S}_1\mathcal{D}_1$ roles as a segmentation-aware denoiser because of the losses on $\mathcal{S}_1$ and $\mathcal{D}_1$. Hence, we train another $\mathcal{S}_1\mathcal{D}_1\mathcal{S}_2$ jointly only with a segmentation loss, denoted as $\mathcal{S}_1\mathcal{D}_1\mathcal{S}_2$-joint, which is a single segmentation network with the same structure as $\mathcal{S}_1\mathcal{D}_1\mathcal{S}_2$. 
The results are on Cityscapes with noise level 50 and shown in Table~\ref{nece_d}. 
$\mathcal{S}_1\mathcal{D}_1\mathcal{S}_2$-joint is inferior to $\mathcal{S}_1$ and $\mathcal{S}_1\mathcal{D}_1\mathcal{S}_2$, which verifies the gain from $\mathcal{S}_1$ to $\mathcal{S}_1\mathcal{D}_1\mathcal{S}_2$ is from the design of segmentation-aware denoiser rather than the more parameters, and the structure of $\mathcal{S}_1\mathcal{D}_1\mathcal{S}_2$ is not suitable for segmentation without the losses on $\mathcal{S}_1$ and $\mathcal{D}_1$.

Besides, we train a wider and deeper $\mathcal{S}_1$ with segmentation loss, denoted as $\mathcal{S}_1+$ in Table~\ref{nece_d}, whose parameter number is comparable to that of $\mathcal{S}_1\mathcal{D}_1\mathcal{S}_2$. The comparison between $\mathcal{S}_1\mathcal{D}_1\mathcal{S}_2$ and $\mathcal{S}_1+$ on segmentation demonstrates that our alternate boosting network outperforms the state-of-the-art segmentation network on noisy images. At the mean time, our method can provide the denoised images, which are unavailable in other segmentation networks.

\begin{table}[!t]
	\small
	\begin{center}
		\begin{tabular}{c|c|c|c}
			\cline{1-4}
			\multicolumn{4}{c}{Cityscapes}\\
			\cline{1-4}
			\multirow{2}*{\#Units}  &\multicolumn{1}{c|}{DDFN~\cite{chen2019real}} & \multicolumn{2}{c}{SDB} \\
			\cline{2-4}
			~&PSNR / SSIM& PSNR / SSIM& Seg. Metrics \\
			\cline{1-4}
			$\times1$&40.83 / 0.9691&40.92 / 0.9695&73.5 / 95.70 / 82.0\\
			$\times2$&40.94 / 0.9695&41.15 / 0.9705&74.6 / 95.73 / 82.9\\
			$\times3$&41.05 / 0.9701&\textbf{41.25} / \textbf{0.9710}&\textbf{75.0} / \textbf{95.81} / \textbf{83.2}\\
			\cline{1-4}
			\multicolumn{4}{c}{OutdoorSeg}\\
			\cline{1-4}
			\multirow{2}*{\#Units}  &\multicolumn{1}{c|}{DDFN~\cite{chen2019real}} & \multicolumn{2}{c}{SDB} \\
			\cline{2-4}
			~&PSNR / SSIM& PSNR / SSIM& Seg. Metrics \\
			\cline{1-4}
			$\times1$& 36.61 / 0.9628&36.64 / 0.9630&75.4 / 88.44 / 85.6\\
			$\times2$& 36.71 / 0.9637 &36.74 / 0.9640&77.1 / 89.25 / 87.3\\
			$\times3$& 36.74 / 0.9640 &\textbf{36.77} / \textbf{0.9643}&\textbf{77.7} / \textbf{89.36} / \textbf{87.4}\\
			\cline{1-4}
		\end{tabular}
	\end{center}
	\caption{Quantitative results of denoising and segmentation in the case of Poisson noise, compared with the denoising method~\cite{chen2019real}. Seg. Metrics include mIoU / Pix. Acc. / Mean Acc..}
	\label{quan_result_poisson}
	\vspace{-0.3cm}
\end{table}

\subsection{Generalization Ability of Noise Distributions}
\label{sec_generalization}
We use the simulated AWGN in the former experiments, but the real-world noisy data may follow a different distribution. 
It is however very difficult to conduct experiments on real noisy images due to the lack of such data with both real-world noise and segmentation labels. Instead, we have conducted a set of experiments using Poisson noise. Different from AWGN, Poisson noise is non-additive and pixel-dependent, whose parameter $\lambda$ on each pixel is different and equal to the corresponding pixel value of clean images. We show the results in Table~\ref{quan_result_poisson}, and conclusions about the alternate boosting are consistent with those on AWGN. From Tables~\ref{quan_result} and~\ref{quan_result_poisson}, we confirm the generalization ability of our method to different noises, including different standard deviations and distributions. 

\section{Conclusions}
We have explored the synergy between image denoising and semantic segmentation by using a holistic deep model, which is a cascade of SDBs. From our experiments, we find denoising and segmentation are both improved significantly with the unit number increasing.  Especially, boosting benefits not only successive denoisers but also successive segmentation modules. In the boosting process, segmentation improves denoising performance and the denoiser helps increase segmentation accuracy.


{\small
\bibliographystyle{ieee_fullname}
\bibliography{refs}
}

\end{document}